\documentclass[conference]{IEEEtran}
\IEEEoverridecommandlockouts

\usepackage{amsmath,amssymb,amsfonts}
\usepackage{algorithmic}
\usepackage{graphicx}
\usepackage{textcomp}
\usepackage{xcolor}

\usepackage{algorithm}
\usepackage{algorithmic}
\usepackage{multirow}

\usepackage{cite}
\usepackage{hyperref}
\usepackage[capitalise]{cleveref}

\usepackage{caption}
\DeclareMathAlphabet\mathbfcal{OMS}{cmsy}{b}{n}

\usepackage{subcaption} 

\def\BibTeX{{\rm B\kern-.05em{\sc i\kern-.025em b}\kern-.08em
    T\kern-.1667em\lower.7ex\hbox{E}\kern-.125emX}}
\begin{document}

\title{Multi-Objective Reinforcement Learning for Power Grid Topology Control
}

\makeatletter
\newcommand{\linebreakand}{%
  \end{@IEEEauthorhalign}
  \hfill\mbox{}\par
  \mbox{}\hfill\begin{@IEEEauthorhalign}
}
\makeatother

\author{\IEEEauthorblockN{Thomas Lautenbacher}
\IEEEauthorblockA{
\textit{50Hertz Transmission GmbH}\\
Berlin, Germany \\
thomasrene.lautenbacher@50Hertz.com\thanks{This research was carried out for an MSc thesis project in Delft University of Technology, supported by the ERASMUS+ program.} 
}
\and
\IEEEauthorblockN{Ali Rajaei}
\IEEEauthorblockA{
\textit{Delft University of Technology}\\
Delft, The Netherlands \\
a.rajaei@tudelft.nl}
\and
\IEEEauthorblockN{Davide Barbieri}
\IEEEauthorblockA{
\textit{TenneT TSO B.V.}\\
Arnhem, The Netherlands \\
davide.barbieri@tennet.eu}
\linebreakand 
\IEEEauthorblockN{Jan Viebahn}
\IEEEauthorblockA{
\textit{TenneT TSO B.V.}\\
Arnhem, The Netherlands \\
jan.viebahn@tennet.eu}
\and
\IEEEauthorblockN{Jochen L. Cremer}
\IEEEauthorblockA{
\textit{Delft University of Technology}\\
Delft, The Netherlands \\
j.l.cremer@tudelft.nl}
}


\maketitle
\thispagestyle{plain}
\pagestyle{plain}

\begin{abstract}
Transmission grid congestion increases as the electrification of various sectors requires transmitting more power. Topology control, through substation reconfiguration, can reduce congestion but its potential remains under-exploited in operations. A challenge is modeling the topology control problem to align well with the objectives and constraints of operators. Addressing this challenge, this paper investigates the application of multi-objective reinforcement learning (MORL) to integrate multiple conflicting objectives for power grid topology control. We develop a MORL approach using deep optimistic linear support (DOL) and multi-objective proximal policy optimization (MOPPO) to generate a set of Pareto-optimal policies that balance objectives such as minimizing line loading, topological deviation, and switching frequency. Initial case studies show that the MORL approach can provide valuable insights into objective trade-offs and improve Pareto front approximation compared to a random search baseline. The generated multi-objective RL policies are $30$\% more successful in preventing grid failure under contingencies and $20$\% more effective when training budget is reduced - compared to the common single objective RL policy.

\end{abstract}



\vspace{-5mm}

\section{Introduction}
The energy transition and the shift toward renewable energy sources are crucial steps for mitigating climate change and ensuring a sustainable energy future. However, this transition poses significant operational challenges for system operators, including congestion management. Transmission network topology control is an under-utilized and non-costly source of flexibility. Adjusting the network topology, such as line switching or modifying busbar connections within substations, can reroute power flows to prevent line overloads and mitigate cascading outages \cite{heidarifar2021optimal,ewerszumrode2024iterative,marot2020learning}. In addition to maintaining continuous electricity supply, power systems must address other objectives, such as minimizing asset wear, reducing operational cost, and mitigating environmental impacts. Achieving these objectives requires a multi-objective approach to decision-making that maintains grid security while addressing other operational objectives \cite{viebahn2024gridoptions}.  

Transmission network topology control problem can be modeled as a mixed-integer non-linear optimization problem which is computationally challenging to approach. The so-called combinatorial explosion of possible topologies and the complex nonlinear nature of power systems \cite{heidarifar2021optimal} makes this problem challenging. To address these challenges, heuristic and expert rule-based approaches, such as in \cite{sogol2023congestion,marot2018expert,hrgovic2024substation,lehna2023compare} are developed to determine corrective topological actions to relieve congestion. However, these approaches do not provide a sequence of control actions and may lead to sub-optimal solutions. To provide sequences of actions, recently researchers explored the use of reinforcement learning (RL) and Artificial Intelligence (AI) more broadly for topological control\cite{Kelly.16.03.2020,marot2020learning, marot2021retrospective, marot2022l2rpnwithtrust}. Studies such as \cite{lan2020ai, yoon2020winning, Subramanian.2021, chauhan2023powrl} explore RL-based approaches, including the deep duelling Q-network (DDQN) initialized with imitation learning \cite{lan2020ai}, the Semi-Markov actor-critic algorithm \cite{yoon2020winning}, the cross-entropy method with importance sampling \cite{Subramanian.2021}, and the proximal policy optimization (PPO) \cite{chauhan2023powrl}. Additionally, \cite{dorfer2022power} develops an AlphaZero-based approach using Monte-Carlo tree search to simulate future outcomes, and guide the agent toward long-term strategies, while \cite{matavalam2022curriculum} presents a curriculum-based approach to improve learning efficiency and stability. Building on these ideas, \cite{Geert2024} combines curriculum learning with tree search to benefit from long-term strategies as well as the efficiency and stability. Some studies focus on addressing the combinatorial explosion of the topology control problem through hierarchical RL \cite{Manczak.Hierarchical} and multi-agent RL \cite{vanderSar.multi-agent}. Furthermore, \cite{hrgovic2024reward} proposes a reward design using multiple metrics to reduce overloads. However, the developed approaches in \cite{marot2018expert,hrgovic2024substation, sogol2023congestion,lehna2023compare,lan2020ai, yoon2020winning, Subramanian.2021,chauhan2023powrl,dorfer2022power,matavalam2022curriculum,Geert2024,Manczak.Hierarchical,vanderSar.multi-agent,hrgovic2024reward} focus on single operational objectives and providing only a single policy. This limits their application to address the trade-offs inherent in the multi-objective nature of power systems and to provide a set of policies for operators to select from.


This paper proposes a multi-objective RL (MORL) approach to address the network topology control problem. Despite previous studies on RL approaches, to the best of the authors’ knowledge, a multi-policy MORL approach for topology control has not been investigated before. To this end, we implement deep optimistic linear support (DOL) and multi-objective PPO (MOPPO) to generate a set of Pareto-optimal policies. Additionally, we develop custom reward functions for different operational objectives, including line loading, topological deviation, and switching frequency. The proposed MORL approach not only shows the trade-offs between these objectives but also provides a set of policies that balance these trade-offs, offering a decision-support approach for system operators. By considering multiple rewards, the approach effectively relieves grid congestion while addressing other operational objectives. 

The rest of the paper is organized as follows. \cref{sec:method} presents the proposed MORL approach and the design of the reward functions. \cref{sec:casestuudies} presents the case studies, investigating the efficiency and robustness of the proposed approach. \cref{sec:dis-conc} provides discussions and concludes the paper.
\vspace{-5mm}
\section{Methodology}
\label{sec:method}
This paper aims to provide a decision-support approach for transmission system operators to perform topological control considering multiple objectives. The proposed approach integrates a multi-objective adaptation of the PPO algorithm \cite{Schulman.20.07.2017,felten2024toolkit} (MOPPO) with deep optimistic linear support \cite{Mossalam.09.10.2016}. To capture different operational objectives, we design custom reward functions that address line loading, topological deviation and switching frequency. The proposed approach considers a multi-policy MORL \cite{Hayes.2022}, which results in a set of optimal solutions rather than a single solution. This allows system operators to better understand the trade-offs among objectives and select the most appropriate policy. \cref{fig:method} depicts the proposed approach. By using DOL as an outer loop method within the MORL approach, the convex coverage set of solutions is constructed iteratively. In each iteration, the DOL generates a new set of weight vectors $\textbf{w}$ and gives one weight vector with the highest priority to MOPPO, that is then trained in the multi-objective environment. MOPPO uses $\textbf{w}$ to account for the multiple rewards $\textbf{R}_t$. After training, the MOPPO is evaluated. If the average value vector over the evaluation episodes $\mathbfcal{V}$, found by the MOPPO, is Pareto optimal, it is added to the convex coverage set (CCS). The detailed methodology is explained in the following.

\begin{figure}[t]
    \centering \includegraphics[width=0.45\textwidth, keepaspectratio=true]{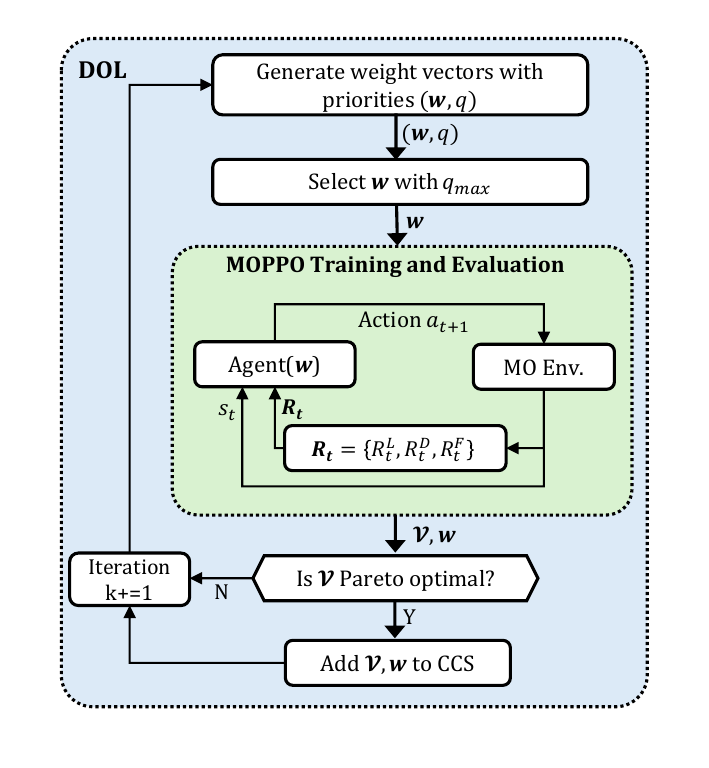}
    \vspace{-0.4cm}
    \caption{Schematic of the proposed MORL approach with deep optimistic linear support.}
    \label{fig:method}
\end{figure}

\subsection{Single-Policy Multi-Objective PPO}
In order to learn on multiple rewards, the agent needs to receive a reward signal for each of the objectives. To this end, we extend the original RL grid environment into a multi-objective environment, allowing the agent to receive a d-dimensional reward vector $\mathbf{r}_{t} \in \mathbb{R}^d$, where $d$ is the number of objectives. In this paper, the rewards reflect the operational objectives of reducing the line loading, decreasing the topological deviation, and reducing the switching frequency, explained in detail in section \ref{se:rewards}.

At the core of the proposed approach is the MOPPO Algorithm \ref{alg:moppo}, which is trained using a reward vector $\mathbf{r}_{t}$ from the environment and a weight vector $\mathbf{w}$, serving as a scalarization, from the DOL to deliver a single policy solution \cite{felten2024toolkit}. The MOPPO algorithm processes the d-dimensional reward vector $\mathbf{r}_{t}$ by using a d-dimensional critic head, resulting in a vectorized value function for each state $s$: 
\begin{equation}
\textbf{V}^\pi(s_t) = \mathbb{E}_\pi \left[ \sum_{t=0}^\infty \gamma^t \mathbf{r}_{t} \,\Big|\, s_t = s \right], \label{eq:valuefunction}
\end{equation}
where $\textbf{V}^\pi(s) \in \mathbb{R}^d$ is the vectorized state-value function per state $s$ under policy \( \pi \), \( \mathbf{r}_{t} \in \mathbb{R}^d \) is the vector of rewards at time \( t\), and \( \gamma \in [0,1) \) is the discount factor.

The vectorized advantages $\mathbf{A}_t \in \mathbb{R}^d $ are then calculated using generalized advantage estimation (GAE) \cite{Schulman.20.07.2017}. These advantages are then scalarized using a weighted sum approach: 

\begin{equation}
\label{eq:adv}
A_t = \mathbf{w}^\top \mathbf{A}_t,
\end{equation}

where \( \mathbf{w} \in \mathbb{R}^d \) is the weight vector representing the scalarization preferences, satisfying \( \sum_{i=1}^n w_i = 1 \) for \( w_i \geq 0 \), and \( A_t \in \mathbb{R} \) is the scalarized advantage. This step incorporates the scalarization function, given by DOL, to train the single-policy MOPPO. The scalarized advantage is then used to calculate the policy loss, which contributes to the total loss. For detailed information on the loss calculation in MOPPO, we refer to  \cite{felten2024toolkit}. The MOPPO produces the final value vector $\textbf{V}$ and the model $m$, containing the neural network weights $\theta$ and the learnt policy $\pi$. 



%
\begin{algorithm}[t]
\caption{Multi-Objective PPO (MOPPO) Training}
\label{alg:moppo}
\begin{algorithmic}[1]
\REQUIRE Multi-Objective MDP, scalarization weight vector $\textbf{w}$, number of update cycles
\STATE initialize MOPPO network $\theta \gets \emptyset$
\FOR{Number of updates cycles}
\STATE initialize replaybuffer($\texttt{B}$) batch $\texttt{B} \gets \emptyset $\\
\STATE $(\textbf{V}_t, \textbf{R}_t)_\texttt{B} \gets\texttt{collect samples} ( \texttt{MOPPO}_{\theta} )$ \\
\COMMENT{Fill replay buffer $\texttt{B}$ by collecting samples from the environment, containing batchsize tuples of vectorized value function and vectorized rewards $(\textbf{V}_t, \textbf{R}_t)_\texttt{B}$}
\STATE $\textbf{A}_t \gets \texttt{compute advantages}((\textbf{V}_t, \textbf{R}_t)\texttt{B})$ 
\STATE $A_t =  \textbf{w}^\top \textbf{A}_t$  \COMMENT{Scalarize advantages \eqref{eq:adv}}
\STATE $\theta \gets \texttt{update}(A_t)$ \COMMENT{updates the neural network}
\ENDFOR
\end{algorithmic}
\end{algorithm}
%
%
\subsection{Reward functions for operational objectives}\label{se:rewards}
\subsubsection{Line Loading Reward}
The \textit{L2RPNReward}, referred to here as \textit{Line Loading Reward}, is the conventional reward used in the L2RPN competition \cite{marot2020learning}, and widely adopted as a default reward in the literature \cite{Manczak.Hierarchical, vanderSar.multi-agent, lehna2023compare, hrgovic2024substation}. This reward emphasizes maintaining adequate thermal loading margins on power lines to ensure grid security.  For each power line \( l \), the thermal loading margin is defined:
\begin{equation}
\text{Margin}_{l,t} = 
\begin{cases}
\displaystyle \frac{\overline{F}_{l} - |F_{l,t}|}{\overline{F}_{l}} & \text{if } |F_{l,t}| \leq \overline{F}_{l} \\[10pt]
0 & \text{if } |F_{l,t}| > \overline{F}_{l}
\end{cases}
\end{equation}
where \( \overline{F}_{l} \) is the thermal limit (ampacity) of line \( l \), and \( |F_{l,t}| \) is the absolute value of the current flow in amps. The line loading reward $R^L_t$ at each time step is calculated as:
\begin{equation}
R^L_t = \sum_{l=1}^{L} \left( \text{Margin}_{l,t} \right)^2
\end{equation}
where, \( L \) represents the total number of power lines. This reward function encourages the agent to keep the power flows within the thermal limits of the lines, penalizing situations where lines are overloaded.

\subsubsection{Topological Deviation}
This reward assigns a penalty based on the degree of deviation of the current grid topology from its initial, default configuration \cite{RTEFrance.2020}. In the default state, all the busbar coupler switches are closed, resulting in a fully meshed configuration and a single electrical node per substation. The reward decreases progressively as the topology deviates from this configuration, encouraging the agent not too deviate too far from the original topology \cite{viebahn2024gridoptions}.

At each time step \( t \), we define the topological deviation for a substation \( i \) as:
\begin{equation}
D_{i,t} = 
\begin{cases} 
1, & \text{if any element in substation } i \\
   & \text{ is assigned to a different bus,} \\ 
0, & \text{otherwise.}
\end{cases}
\end{equation}

The total topological deviation at time \( t \) is :
\begin{equation}
D_{t} = \sum_{i=1}^{N} D_{i,t}
\end{equation}
where \( N \) denotes the total number of substations.

The Deviation reward at time \( t \) is:
\begin{equation}
R^D_t = 
\begin{cases} 
\textcolor{blue}{r_{\text{default}}^\text{D}}, & \text{if } D_t = 0, \\ 
\textcolor{blue}{r_{\text{low}}^\text{D}}, & \text{if } D_t \leq d_{\text{threshold}}, \\ 
\textcolor{blue}{r_{\text{high}}^\text{D}}, & \text{otherwise,}
\end{cases}
\end{equation}
where \( \textcolor{blue}{r_{\text{default}}^\textbf{D}} \) is a positive reward indicating the maintenance of the default state, \( \textcolor{blue}{r_{\text{\text{low}}}^\text{D}} \) is a small penalty for minor deviations from the default configuration, \( \textcolor{blue}{r_{\text{high}}^\text{D}} \) is a larger penalty for significant deviations, and \( d_{\text{threshold}} \) is the predefined threshold for minor deviations.

Transmission networks operate more securely and more resilient to disturbances in a fully meshed configuration with minimal or no changes to the topology \cite{marot2020learning}. Consequently, the piecewise linear design of the reward functions incentivizes the agent not to minimize deviation  from the default configuration by imposing disproportionately higher penalties for significant alterations to the grid topology.

\subsubsection{Switching Frequency}
This reward penalizes the number of switching actions within a specified time interval, aiming to discourage instability from frequent adjustments \cite{viebahn2024gridoptions}.
To prevent the reward from exploding, we consider only the accumulated switching actions within one time interval. Thus, the total possible episode is divided into intervals $\{\mathcal{T}_m\}$ (e.g., timesteps within one hour or within one day). At each RL time step $t$, the cumulative switching actions $F_t$ in interval $\{\mathcal{T}_m\}$ is computed as:
\begin{equation}
F_t = \sum_{t \in \mathcal{T}_m} \text{switching actions in interval $m$ up to $t$}
\end{equation}
The switching frequency reward is defined as:
\begin{equation}
R^F_t = 
\begin{cases}
\textcolor{blue}{r_{\text{default}}^\text{F}}, & \text{if } F_t = 0, \\
\textcolor{blue}{r_{\text{low}}^\text{F}}, & \text{if } F_t \leq F_{\text{low}}^\text{th}, \\
\textcolor{blue}{r_{\text{high}}^\text{F}},& \textcolor{blue}{\text{otherwise}}
\end{cases}
\end{equation}
where \( \textcolor{blue}{r_{\text{default}}^\text{F}} \) is a baseline reward indicating no switching actions, \( \textcolor{blue}{r_{\text{low}^\text{F}}} \) is a penalty for low switching frequency, \( \textcolor{blue}{r_{\text{high}}^\text{F}} \) is a penalty for high switching frequency, \textcolor{blue}{and \( F_{\text{low}}^\text{th} \) is the predefined threshold for low and high switching frequencies, respectively.}
\textcolor{blue}{ $R^F_t$ focuses on the operational interactions in the grid e.g. material wear off and effort for operators, whereas $R^D_t$ is directed to operational robustness of the power grid \cite{viebahn2024gridoptions}.}

\subsection{Deep Optimistic Linear Support}

\begin{algorithm}[t]
\caption{Deep Optimistic Linear Support (DOL)}\label{alg:dol}
\begin{algorithmic}[1]
\REQUIRE single-policy RL algorithm \texttt{MO PPO}, maximum number of iterations $k^{max}$, reuse option (\texttt{no reuse}, \texttt{full reuse}, \texttt{partial reuse})
\STATE Initialize partial CCS $\Omega^s \gets \emptyset$
\STATE Initialize set of visited weights $W \gets \emptyset$
\STATE Initialize priority queue $Q \gets \emptyset$
\STATE Initialize model repository $\text{Models} \gets \emptyset$
\STATE Initialize iteration $k$ as 0. 
\FORALL{extremum weight $w_e$ of the weight simplex}
    \STATE Add $(w_e, \infty)$ to $Q$ \COMMENT{Add extrema with infinite priority}\label{alg:line-inf}
\ENDFOR
\WHILE{$Q$ is not empty \AND $k<k^{max}$} \label{alg:k}
    \STATE Pop $w$ from $Q$ with the highest priority $q_{\text{max}}$ \label{alg:pop_q}
    \IF{reuse option is \texttt{no reuse} \OR $\text{Models}$ is empty}\label{alg-reuse0}
        \STATE Initialize model parameters $\theta$ randomly
    \ELSE
        \STATE Find the closest weight $w_{\text{closest}}$ in $W$ to $w$
        \STATE Initialize model parameters $\theta$ with parameters from $\text{Models}[w_{\text{closest}}]$
        \IF{reuse option is \texttt{partial reuse}}
            \STATE Randomly reinitialize the last layer of the model
        \ENDIF
    \ENDIF\label{alg-reuse1}
    \STATE $\theta_{\text{new}} \gets \texttt{MO PPO}(m, w, \theta)$ \COMMENT{Train RL algorithm with weight $w$}
    \STATE $\mathbfcal{V} \gets \texttt{MO PPO}(m, w, \theta_{\text{new}})$ \COMMENT{Evaluate RL algorithm with weight $w$} \label{alg:evaluate}
    
    \STATE $W \gets W \cup \{ w \}$
    \IF{there exists $w'$ such that $w' \cdot \mathbfcal{V} > \max\limits_{U \in \Omega^s} w' \cdot U$} \label{alg:improvement}
        \STATE Remove corner weights made obsolete by $\mathbfcal{V}$ from $Q$, store them in $W_{\text{del}}$ \label{alg:remove}
        \STATE $W_{\text{del}} \gets W_{\text{del}} \cup \{ w \}$
        \STATE $\Omega^s \gets \Omega^s \cup \{ \mathbfcal{V} \}$
        \STATE Remove vectors from $\Omega^s$ that are no longer optimal after adding $\mathbfcal{V}$
        \STATE $\text{Models}[w] \gets \theta_{\text{new}}$
        \STATE $(W_{\text{new}}, q_{\text{new}}) \gets \texttt{newCornerWeights}(\Omega^s, V)$\label{alg:new-corner}
    \ENDIF \\
    $k+=1$
\ENDWHILE
\RETURN $\Omega^s$ and the models in $\text{Models}$ \COMMENT{models correspond to the policies implicitly integrated in the neural network weights}
\end{algorithmic}
\end{algorithm}

To generate a set of policies, we employ the deep optimistic linear support (DOL) algorithm to iteratively construct the optimal solution set of policies \cite{Mossalam.09.10.2016}. DOL functions as an outer-loop MORL approach \cite{Hayes.2022}, leveraging optimistic linear support (OLS) \cite{roijers2016multi} to iteratively construct the convex coverage set (CCS) of a multi-objective problem by generating scalarization functions in the form of weight vectors $\mathbf{w}$ and updating the solution set $\Omega^s$ accordingly \cite{Hayes.2022}. 
\textcolor{blue}{This outerloop approach is choosen, as to incorporate already existing research in single objective RL algorithms which solve the topology control problem}\cite{Mossalam.09.10.2016}. 
Algorithm \ref{alg:dol} provides details of the proposed DOL approach. In the first iteration, DOL assigns the extrema weights to the queue with infinite priority, ensuring these weights are processed first by the MOPPO algorithm (line \ref{alg:line-inf}). 
For each new iteration of the DOL algorithm, the weight vector with the highest priority $q_{max}$ (expected highest improvement) will be given to the MOPPO (line \ref{alg:pop_q}). The MOPPO is trained using $\mathbf{w}$, described in Algorithm \ref{alg:moppo}. The trained agent containing $\theta_{new}$ is evaluated and produces $\mathbfcal{V}$ as the average value vector across the evaluation episodes (line \ref{alg:evaluate}). $\mathbfcal{V}$ is used to evaluate if the trained model contributes to the CCS. $\mathbfcal{V}$ is the sampled vectorized value-estimation $\mathbf{V}(s)$ over the evaluation episodes, based on the received rewards; hence, $\mathbfcal{V}$ does not necessarily reflect the vectorized state dependent value function $\mathbf{V}(s)$ (\ref{eq:valuefunction}).

If the new value vector improves the coverage set $\Omega^s$, obsolete corner weights are deleted, the new value vector is added to the set, the model is saved and new corner weights are calculated (line \ref{alg:improvement}-\ref{alg:new-corner}). To determine next weight vectors, corner weights are identified as the weights where the piecewise-linear convex (PWLC) surface of $\mathbfcal{V}(w)$ changes slope (line \ref{alg:new-corner}). Specifically, these corner weights are the vertices of the polyhedral subspace above $\mathbfcal{V}(w)$. The priority of the new corner weights is calculated based on their distance to the assumed optimistic upper bound of the CCS. Details on the $\texttt{newCornerWeights}$ are provided in \cite{roijers2016multi}.
The original stopping criterion, which depended on a minimum improvement in the CCS \cite{Mossalam.09.10.2016}, is replaced with a predefined maximum number of iterations, $k^{\text{max}}$ (line \ref{alg:k}). This allows for an initial estimate of the training effort by fixing the number of iterations in advance. 

DOL offers the possibility of reusing model parameters from the nearest model (lines \ref{alg-reuse0}-\ref{alg-reuse1}). The neural network weights $\theta$ from the model with the closet $\textbf{w}$ are used to initialize the next MOPPO. For a detailed view on DOL and OLS, we refer to \cite{Mossalam.09.10.2016} and \cite{roijers2016multi}, respectively.

\subsection{Policy Selection}
Despite the multi-objective nature of the problem, the ultimate goal of the system operator is maintain secure system operation as long as possible \cite{viebahn2024gridoptions}. 
To this end, Algorithm \ref{alg:selection} is proposed to identify the best-performing policy as a recommendation for the system operator. We assess policy quality using a metric independent of the rewards, referred to as \textit{Episode Duration} (\(E\)) \cite{RTEFrance.2020}. $E$ measures how long the agent can prevent the power system from premature grid failure.
After generating the complete set of policies, we select the best-performing policy $(\pi^{MO})$ for each seed run based on (\( E \)). Policies trained solely on extreme weights are excluded from consideration, as the focus is on selecting policies optimized for multiple rewards.

\begin{algorithm}[t]
\caption{Selection Process for Best Performing Policy}\label{alg:selection}
\begin{algorithmic}[1]
\REQUIRE Set of policies $\{\pi_1, \pi_2, \ldots, \pi_n\}$, set of episode durations $\{E_1, E_2, \ldots, E_n\}$, extrema weights $\{w_e\}$
\STATE Initialize best policy $\pi^{MO} \gets \emptyset$
\STATE Initialize maximum episode duration $E^{\text{max}} \gets 0$
\FORALL{$i \in \{1, 2, \ldots, n\}$}
\STATE Retrieve policy $\pi_i$ and corresponding episode duration $E_i$
    \IF{$\pi_i$ is not trained on extrema weights $\{w_e\}$ \AND $E_i > E^{\text{max}}$}
        \STATE $E^{\text{max}} \gets E_i$
        \STATE $\pi^{MO} \gets \pi_i$
    \ENDIF
\ENDFOR
\RETURN $\pi^{MO}$
\end{algorithmic}
\end{algorithm}

\section{Case Studies}
\label{sec:casestuudies}

\subsection{Settings}

\begin{figure}[t]
    \centering\includegraphics[width=0.6\linewidth, keepaspectratio=true,trim={0cm 6cm 17cm 0cm},clip]{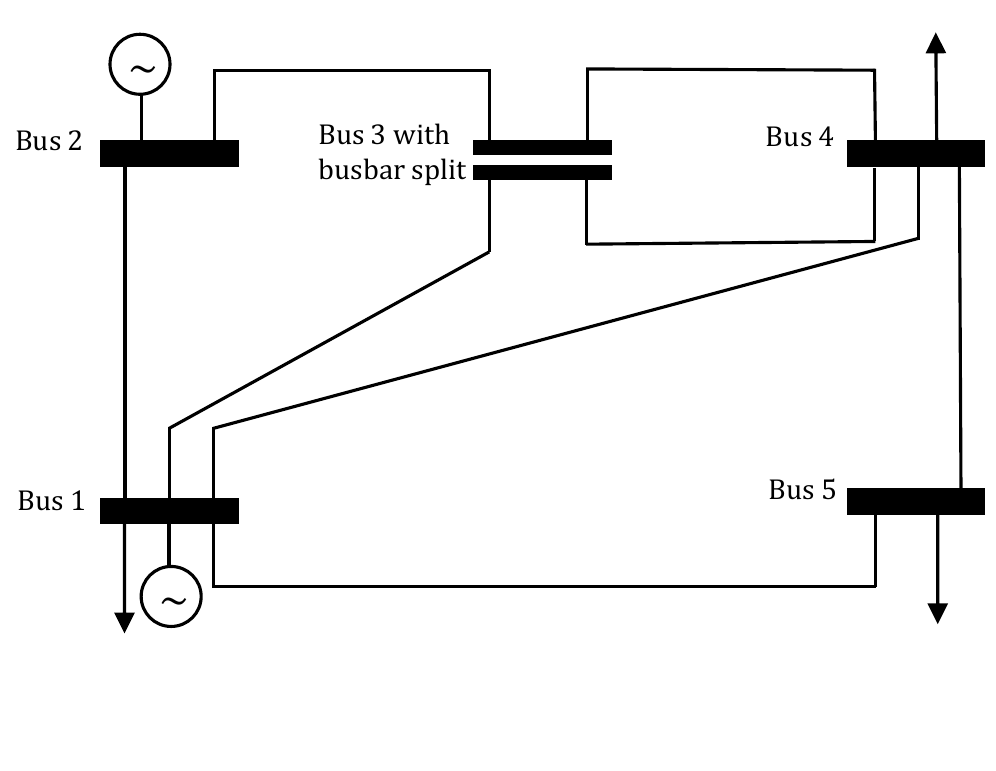}
    \vspace{-1cm}
    \caption{Schematic of the RTE $5$-bus system with busbar splitting on substation 3.}
    \vspace{-5mm}
    \label{fig:rte5bus}
\end{figure}

All case studies are performed on the RTE 5-bus system in the Grid2Op environment \cite{RTEFrance.2020}, providing  initial insights into a MORL approach for topology control. \textcolor{blue}{As this is the first study on the inquiry of MORL on topology control we focus on the RTE 5-bus system.} Figure \ref{fig:rte5bus} shows the 5-bus system. Each environment scenario lasts a week with a 5-min resolution (2016 time steps), which corresponds to the maximum episode duration ($E$). \textcolor{blue}{As only little scenarios are available, we} use 16 scenarios for training, 2 scenarios for validation, and 2 scenarios for testing. \textcolor{blue}{The data used is available, when initializing the respective environment from Grid2Op \cite{RTEFrance.2020}.} The experiments are performed on up to 20 random seeds initializing the environment. The PPO neural network architecture consists of 2 fully connected layers with 64-dimensional hidden features. Training is performed using the Adam optimizer with a learning rate of $5 \times 10^{-4}$ and a batch size of 512. The MOPPO algorithm assumes 4 update cycles. \textcolor{blue}{Initial hyperparameters are derived from \cite{Manczak.Hierarchical} and optimized using empirical experience.} In the case study on robustness to contingencies (\cref{sec3:robust}), an adversarial agent is considered to simulate N-1 contingency states by randomly targeting power lines. Additionally, a set of common expert rules are considered to improve the performance and ensure safety, as detailed in \cite{Subramanian.2021, lehna2023compare}. All computations  are performed using DelftBlue's supercomputer, equipped with Intel XEON E5-6248R 24C 3.0GHz CPU cores \cite{DHPC2024}. We use Grid2Op 1.10, LightSim2Grid 0.8, pandapower 2.14, gymnasium 0.29, mo-gymnasium 1.1 and the morl-baselines 1.0 package. The code for this study is publicly available in \cite{github}.

In Section \ref{PFapprox}, 
we compare our approach to a random sampling (RS) benchmark, which replaces the DOL component by randomly selecting weight vectors from a uniform distribution. These randomly selected weights are then provided to the MOPPO, following the same process as in the DOL-based approach. To enhance this baseline, we incorporate the extrema weights \cite{roijers2016multi}, as exploring these weights is expected to yield significant gains in the objective space.


\subsection{Pareto Front Approximation}\label{PFapprox}

\cref{tab:hypervolume_sparsity_mean} presents the results for hypervolume, sparsity and inverted generational distance (IGD) of the proposed DOL approach and RS. 
The hypervolume metric evaluates the spread and distribution of the solution space, the sparsity metric quantifies the density of the solution set, and IGD measures how accurately the generated solution set approximates the true Pareto front \cite{Hayes.2022}. The DOL and RS achieve similar mean hypervolume, with DOL slightly outperforming RS. However, DOL shows 50\% lower mean sparsity compared to RS, indicating a much denser coverage of the Pareto front, more suitable for a decision support tool. Additionally, DOL exhibits a substantial reduction in IGD by 60\%, indicating a better approximation of the assumed true convex coverage set.

\begin{table}
\renewcommand{\arraystretch}{1.3}
    \centering
    \resizebox{0.45\textwidth}{!}{
    \begin{tabular}{c|c c c c c c}
        \hline
        \multirow{2}{*}{Approach} & HV & HV & Spa & Spa & IGD & IGD \\
         & Mean & Std & Mean & Std & Mean & Std \\
        \hline
        DOL & 44.62 & 27.32 & 0.11 & 0.04 & 0.84 & 0.20 \\
        RS  & 37.35 & 31.67 & 0.22 & 0.06 & 2.22 & 0.12 \\
        \hline
    \end{tabular}}
    \caption{Hypervolume, sparsity and inverted generational distance for DOL and RS Methods.}
    \vspace{-3mm}
    \label{tab:hypervolume_sparsity_mean}
\end{table}

\begin{figure}[t]
    \centering
    \includegraphics[width=0.8\linewidth]{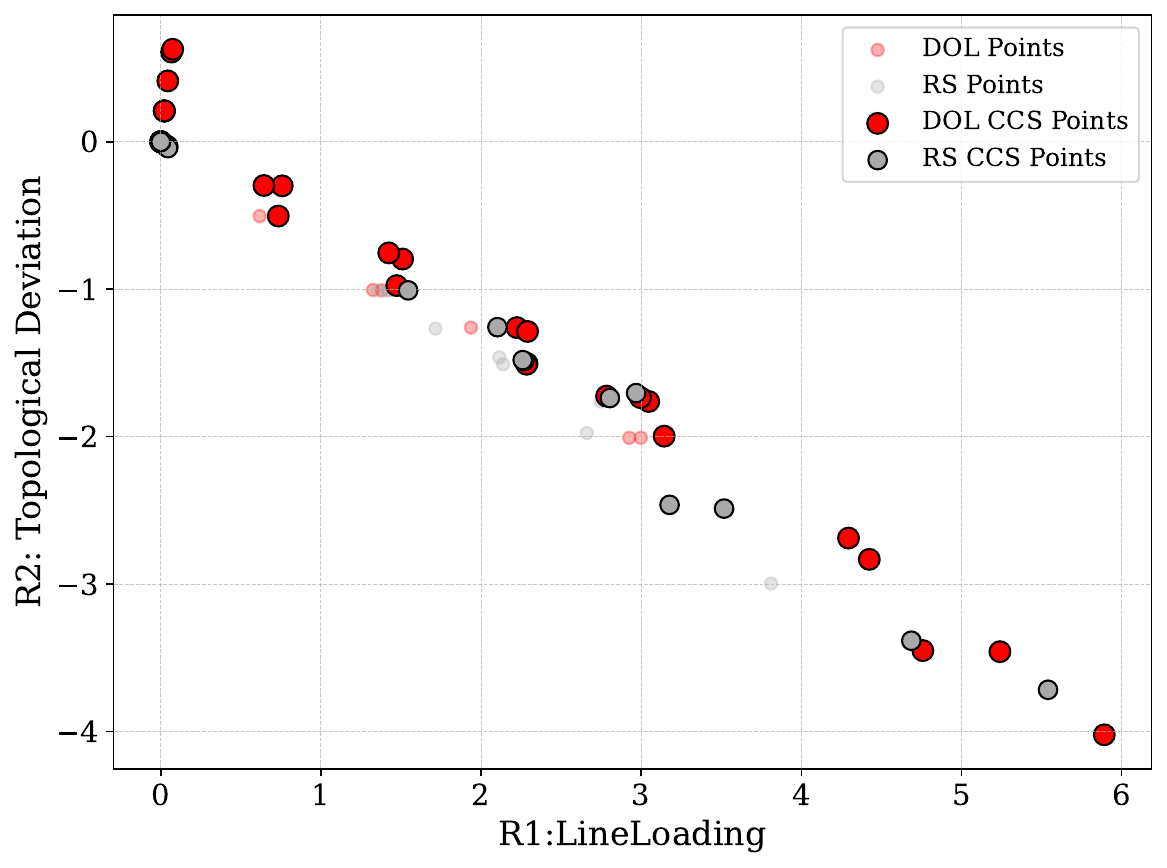}
    \vspace{-1mm}
    \caption{2D Projection of Super CCS for line loading reward vs topological deviation Reward.}
    \vspace{-3mm}
    \label{fig:line_loading_vs_topological_depth}
\end{figure}

\begin{figure}[t]
    \centering
    \includegraphics[width=0.8\linewidth]{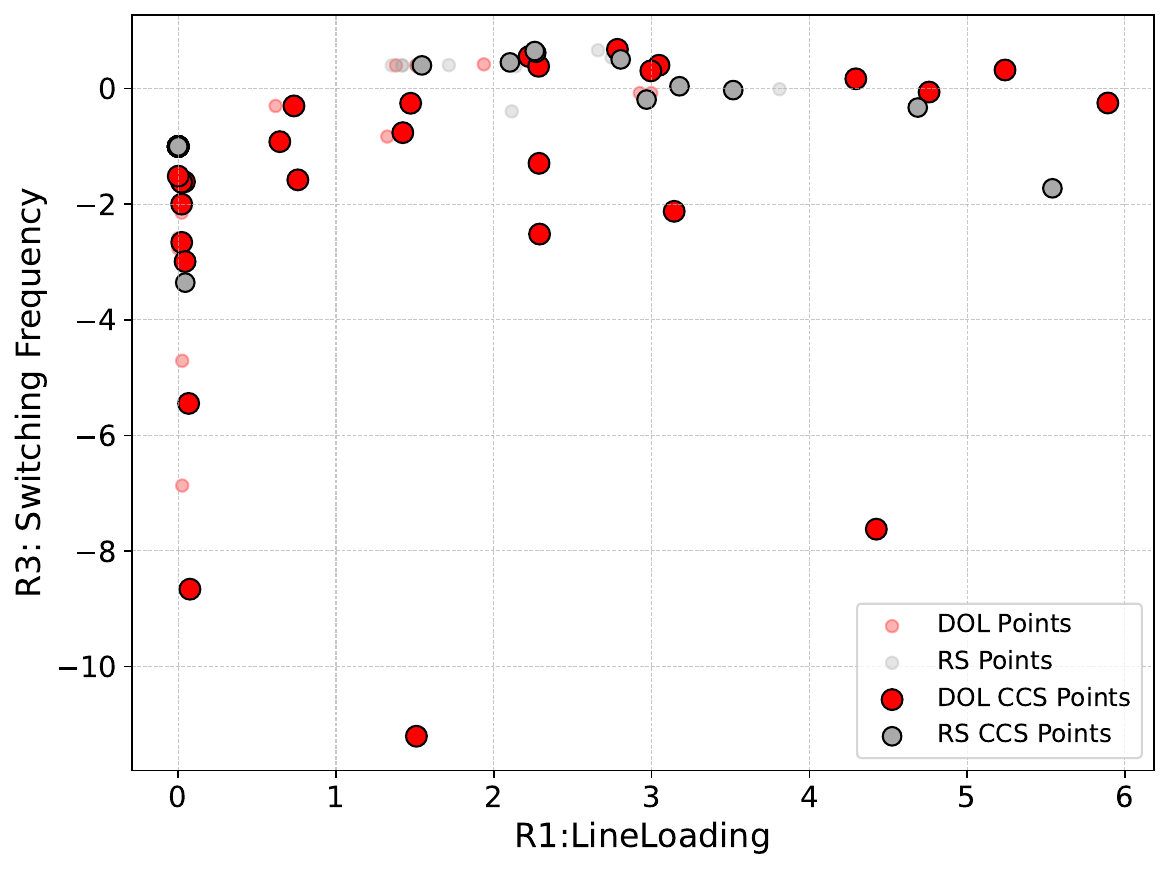}
    \vspace{-1mm}
    \caption{2D Projection of Super CCS for line loading reward vs switching frequency reward.}
    \vspace{-3mm}
    \label{fig:line_loading_vs_switching_frequency}
\end{figure}

\begin{figure}[t]
    \centering
    \includegraphics[width=0.8\linewidth]{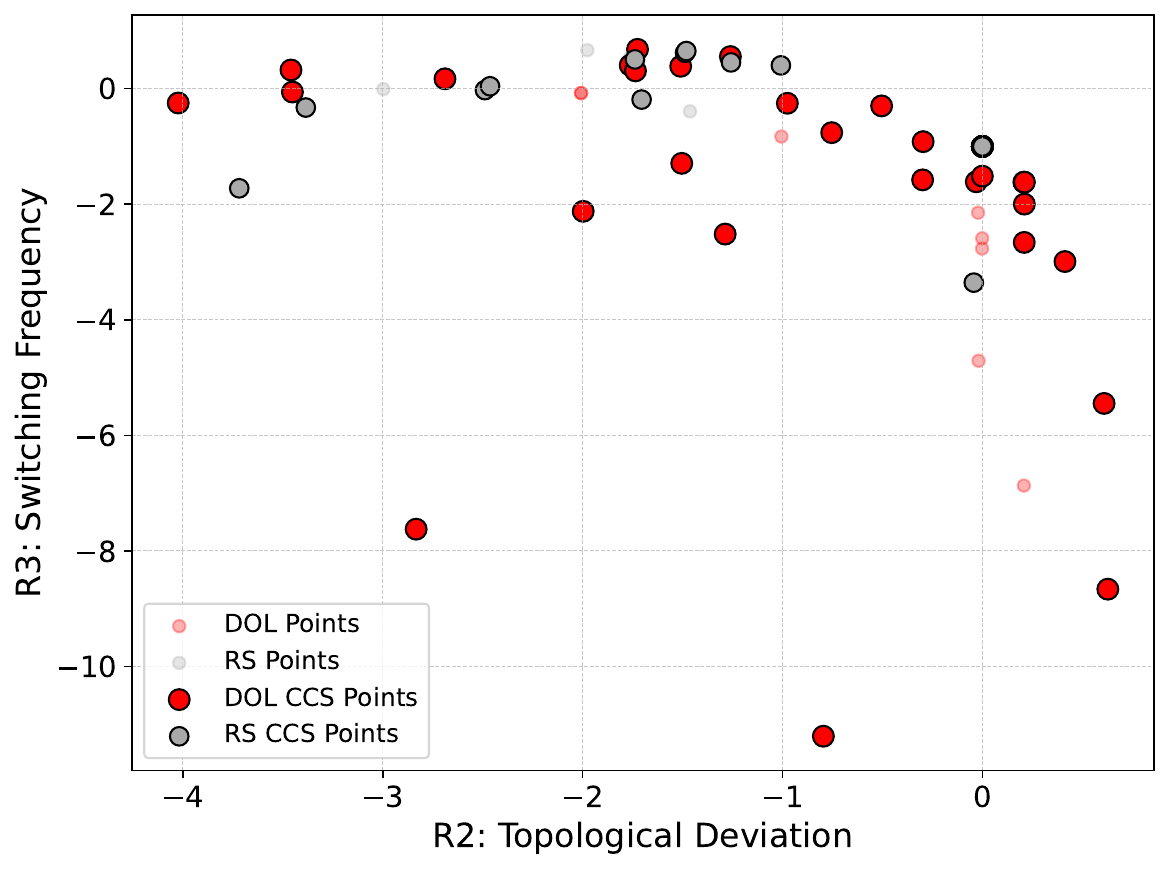}
    \vspace{-1mm}
    \caption{2D Projection of Super CCS for topological deviation reward vs switching frequency reward}
    \vspace{-3mm}
    \label{fig:topological_depth_vs_switching_frequency}
\end{figure}

\cref{fig:line_loading_vs_topological_depth,fig:line_loading_vs_switching_frequency,fig:topological_depth_vs_switching_frequency} show the 2D projections of the super CCS for DOL and RS runs across five seeds into 2-dimensional reward spaces. The super CCS is constructed as the convex set over all generated solution sets across all seeds and both DOL and RS generated solutions. The Super CCS here serves as an indicator for the assumed true CCS. The DOL generates more points compared to the RS benchmark that contribute to the formation of the super CCS. Assuming the super CCS reflects the true trade-offs in the objective space, we can conclude that DOL more closely approximates these trade-offs.

\cref{fig:line_loading_vs_topological_depth,fig:line_loading_vs_switching_frequency,fig:topological_depth_vs_switching_frequency} illustrates key trade-offs among the objectives. In \cref{fig:line_loading_vs_topological_depth}, a clear conflict is observed between $R^L$ and $R^D$, where reducing topological deviation often results in lower line loading. This indicates that changes in topology are sometimes necessary to maintain grid security, trading off topological deviation for improved line loading. 

In \cref{fig:line_loading_vs_switching_frequency}, low $R^L$ corresponds to low $R^S$, indicating that minimal switching interaction allows the grid to operate securely. However, some switching is necessary to further increase $R^L$. The trade-off solutions in the top middle of \cref{fig:line_loading_vs_switching_frequency} may appeal to power system operators who seek an RL policy balancing low switching frequency with low line loading. 

In \cref{fig:topological_depth_vs_switching_frequency}, $R^D$ and $R^S$ exhibit a trade-off, as lower switching frequency can lead to higher topological deviation. This occurs because deviations in topology persist longer without switching actions, leaving the grid in a deviated state for extended periods.
\vspace{-1mm}
\subsection{Robustness to N-1 Contingencies}
\label{sec3:robust}

This case study investigates the robustness of MO policies that are trained on multiple rewards compared to SO policies under N-1 contingency states. The contingency states are generated by an adversarial attacker, which disconnects power lines at random. We use the same settings for the adversarial attacks as in \cite{Manczak.Hierarchical}. The multi-objective policies (MO policies) are trained on $R^L$, $R^D$, and $R^F$ rewards, while the SO policy is trained on the common $R^L$ reward. The MO policies are selected based on Algorithm \ref{alg:selection} considering the Episode Duration metric $E$. The following scenarios are considered:
\begin{itemize}
    \item No Contingencies: The environment does not include any unplanned contingencies (baseline).
    \item Moderately Frequent Contingencies: line disconnection randomly at maximum twice a day.
    \item Highly Frequent Contingencies: line disconnection randomly at maximum four times a day.
\end{itemize}

\begin{table}[t]
    \renewcommand{\arraystretch}{1.3}
    \centering
    \begin{tabular}{c c c c c }
        \hline
        &  & N-1 Contingency Frequency  & \\
        & No & Moderate & High \\
        \hline
        $E^{MO}$(\%) & 94.83 & 97.68 & 90.33 \\
        $E^{SO}$(\%) & 82.66 & 58.61 & 66.47\\
        $\Delta E$ (\%) & 10.17 & 39.07 & 23.86 \\
        \hline
    \end{tabular}
    \caption{Comparison of episode duration ($E$) for multi objective and single objective policies under N-1 contingencies.}
    \label{tab:average_episode_duration}
\end{table}

Table \ref{tab:average_episode_duration} compares the mean episode duration ($E$) and the improved episode duration ($\Delta E$) normalized by the the total number of possible steps for the MO and SO policies. The results show that MO RL policies achieve a higher average episode duration. For instance, for highly frequent contingencies, the MO Policies achieve 23.86\% increase in episode duration compared to SO policies. In the setting with moderately frequent contingencies, the MO Policies outperform the SO policies by almost 40\%. By learning to reduce the topological deviation and to reduce the switching frequency, agents trained with MO policies develop more robust strategies, which perform better under contingencies.  
\subsection{Efficient Training}
This case study investigates the training efficiency of MO and SO policies when computational resources are limited. Similar to the previous case study, we evaluate performance using the average episode duration and select the best MO policies according to Algorithm \ref{alg:selection}. \cref{tab:training} compares MO and SO policies considering the following training scenarios:
\begin{itemize}
    \item Full Training Budget: The agent is trained with the default number of interactions (2048 training samples, 4 update cycles).
    \item Moderate Training Budget: The agent is trained on 75\% of the training (1536 training samples, 3 update cycles).
    \item Low Training Budget: The agent is trained on 50\% of the training (1024 training samples, 2 update cycles). 
\end{itemize}

\begin{table}[t]
    \centering
    \renewcommand{\arraystretch}{1.3}
    \centering
    
    \begin{tabular}{c c c c c }
        \hline
        & & Training Budget & \\
        & Low & Moderate & Full \\
        \hline
        $E^{MO}$ (\%) &  95.27 & 90.44 & 94.83 \\
        $E^{SO}$ (\%) & 73.39  & 84.95 &  82.66\\
        $\Delta E$  (\%) & 21.88  & 5.49 & 10.17 \\
        \hline
    \end{tabular}
    \caption{Comparison of episode duration ($E$) for multi objective and single objective policies under constrained training.}
    \vspace{-3mm}
    \label{tab:training}
\end{table}

\cref{tab:training} shows that MO policies outperform SO policies with fewer training iterations. Notably, in the low training scenario, MO policies achieve a 21.88\% higher episode duration on average. By focusing on reducing switching frequency and maintaining proximity to the original topology early in training, MO policies develop effective strategies at an earlier stage. As a result, MO policies provide faster and more efficient learning, a critical advantage as the computational complexity of larger grids increases exponentially.

\section{Discussion and Conclusion}
\label{sec:dis-conc}
This paper presents the first investigation into multi-objective reinforcement learning (MORL) for power grid topology control. We demonstrated trade-offs exist among conflicting operational objectives in the underlying problem of topological control. From analyzing the initial case studies, we conclude the approach seems promising to alleviate challenges in modeling this topological control problem. In other words, the underlying problem being multi-objective can be approached through learning in a more principled way. The initial case studies show that the proposed DOL approach generates higher-quality solution sets, with higher density and closer approximation of the Pareto frontier, compared to random sampling, thereby offering enhanced decision support and a more comprehensive set of policies for operators to select from. Additionally, by simultaneously reducing topological deviation, switching frequency and line loading, multi-objective policies achieve 24\% higher average episode duration under high-contingency scenarios, compared to single-objective policies. The results also show that multi-objective policies improve training efficiency; when using $50$\% of the training steps, MO policies achieve a 22\% better episode duration compared to single-objective policies. However, some limitations of the study can be noted. The 5-bus system used in this research does not fully reflect the complexity of real-world power systems. Future work should adapt the proposed approach to a larger grid to investigate its practical applicability. Additionally, this study focuses on a limited set of operational objectives, including line loading and switching frequency. Future research should explore other objectives, such as operational cost, and environmental impacts. Moreover, expanding MORL for topology control to include both topological actions and generator re-dispatch should be explored in future research.

\vspace{-1mm}
\section*{Acknowledgment}
\vspace{-2mm}
This work was supported through the AI-EFFECT project (Grant Agreement No 101172952) and AI4REALNET (Grant Agreement No 101119527) ); funded under the European Union’s Horizon Europe Research and Innovation program. However, views and opinions expressed are those of the author(s) only and do not necessarily reflect those of the European Union.


\ifCLASSOPTIONcaptionsoff
  \newpage
\fi

\newcommand{\BIBdecl}{\setlength{\itemsep}{-0.2em}}

\bibliographystyle{IEEEtran}
\bibliography{Library}

\end{document}